\documentclass[preprint,authoryear,12pt]{elsarticle}
\usepackage{graphicx}
\usepackage{amssymb}
\usepackage[top=2.5cm,left=2.5cm,right=2.5cm,bottom=2.5cm]{geometry}% by courtesy of Mico
\usepackage{csvsimple}
\usepackage{multirow}
\usepackage{lipsum}
\usepackage{afterpage}

\usepackage{etoolbox}
\robustify\bfseries
\usepackage{subcaption}

\usepackage{eurosym}
\usepackage{siunitx}

    \DeclareSIUnit\eur{\officialeuro}
    \DeclareSIUnit\M{M}
    \DeclareSIUnit\k{k}

  \def\sym#1{\ifmmode^{#1}\else\(^{#1}\)\fi}

\usepackage{setspace} 

\usepackage{csquotes}
\usepackage{amsmath}
\usepackage{dsfont}
\usepackage{bm}

\usepackage{xspace}
	\newcommand\ie{i.\,e.\xspace}
	\newcommand\eg{e.\,g.\xspace}

	\newcommand\cf{cf.\xspace}

\usepackage[amsmath,hyperref,framed]{ntheorem} %,thmmarks <--endmark
  \theoremstyle{plain}

  \theoremstyle{nonumberplain}
    \theoremseparator{.}
    \theoremheaderfont{\bfseries}
    \theorembodyfont{\normalfont}
    \theoremsymbol{$\blacksquare$}
      \RequirePackage{amssymb}

      \makeatletter
    \let\copy@theorem@headerfont=\theorem@headerfont
    \newcommand{\my@theorem@headerfont}{%
        \boldmath\copy@theorem@headerfont\unboldmath
      }
    \let\theorem@headerfont=\my@theorem@headerfont
      \makeatother
   
\theoremstyle{nonumberplain}
\theoremseparator{.}
\newcommand{\argmax}{\operatornamewithlimits{arg \, max}}

\usepackage[%
  sort&compress
]{cleveref}
\usepackage{url}

\usepackage{isomath}

\usepackage{algorithm}% http://ctan.org/pkg/algorithm
\usepackage{algpseudocode}% http://ctan.org/pkg/algorithmicx

\usepackage[usenames,dvipsnames]{xcolor}

\usepackage{colortbl}
\usepackage{tabularx}
\usepackage{booktabs}
\usepackage{collcell}

\usepackage{ragged2e}
  
\newcommand{\PreserveBackslash}[1]{\let\temp=\\#1\let\\=\temp}
\newcolumntype{v}[1]{>{\PreserveBackslash\RaggedRight\hspace{0pt}}p{#1}}

  \usepackage{adjustbox}
\newcolumntype{Q}[2]{%
    >{\adjustbox{angle=#1,lap=\width-(#2)}\bgroup}%
    l%
    <{\egroup}%
}
\newcommand{\mcellt}[2][c]{%
  \begin{tabular}[t]{@{}#1@{}}#2\end{tabular}}
\newcommand{\lcellt}[2][l]{%
  \begin{tabular}[t]{@{}#1@{}}#2\end{tabular}}

\usepackage[
    colorinlistoftodos,
    textsize=footnotesize,
        ]{todonotes}

\makeatletter
    \renewcommand{\fps@figure}{htb}       
    \renewcommand{\fps@table}{htbp}        
\makeatother 

\usepackage{float}
\usepackage{rotating}

\usepackage{placeins}

\journal{Decision Support Systems}

\begin{document}
\begin{frontmatter}

\title{Deep learning for affective computing:  text-based emotion recognition in decision support}

\author[ETH]{Bernhard Kratzwald}
\ead{bkratzwald@ethz.ch}

\author[NII]{Suzana Ili\'{c}\corref{cor1}}
\ead{suzana.ilic.nii@gmail.com}

\author[ETH]{Mathias Kraus}
\ead{mathiaskraus@ethz.ch}

\author[ETH]{Stefan Feuerriegel}
\ead{sfeuerriegel@ethz.ch}

\author[NII]{Helmut Prendinger}
\ead{helmut@nii.ac.jp}

\address[ETH]{ETH Zurich, Weinbergstr. 56/58, 8092 Zurich, Switzerland}

\address[NII]{National Institute of Informatics, 2-1-2 Hitotsubashi, 101-8430 Tokyo, Japan}

\cortext[cor1]{Corresponding author.}

\begin{abstract}
Emotions widely affect human decision-making. This fact is taken into account by affective computing with the goal of tailoring decision support to the emotional states of individuals. However, the accurate recognition of emotions within narrative documents presents a challenging undertaking due to the complexity and ambiguity of language. Performance improvements can be achieved through deep learning; yet, as demonstrated in this paper, the specific nature of this task requires the customization of recurrent neural networks with regard to bidirectional processing, dropout layers as a means of regularization, and weighted loss functions. In addition, we propose \emph{sent2affect}, a tailored form of transfer learning for affective computing: here the network is pre-trained for a different task (i.e. sentiment analysis), while the output layer is subsequently tuned to the task of emotion recognition. The resulting performance is evaluated in a holistic setting across 6 benchmark datasets, where we find that both recurrent neural networks and transfer learning consistently outperform traditional machine learning. Altogether, the findings have considerable implications for the use of affective computing. 
\end{abstract}


\end{frontmatter}

\section{Introduction}

Emotions drive the ubiquitous decision-making of humans in their everyday lives~\citep{Oatley2011, Greene2002, Schwarz2000}. Furthermore, emotional states can implicitly affect human communication, attention, and the personal ability to memorize information~\citep{Derakshan2010, Dolan2002}. While the recognition and interpretation of emotional states often comes naturally to humans, these tasks pose severe challenges to computational routines~\citep[\eg,][]{Poria2017, Tausczik2010a}. As such, the term \emph{affective computing} refers to techniques for detecting, recognizing, and predicting human emotions (\eg, joy, anger, sadness, trust, surprise, anticipation) with the goal of adapting computational systems to these states~\citep{Picard1997}. The resulting computer systems are not only capable of exhibit empathy~\citep{Picard1995} but can also provide decision support tailored to the emotional state of individuals. 

Emotional information is conveyed through a multiplicity of physical and physiological characteristics. Examples of such indicators include vital signs such as heart rate, muscle activity or sweat production on the surface of the skin~\citep[\eg,][]{Lux2015,Tao2005}. A different stream of research tries to infer emotions from the content and its mode of communication. These approaches to affective computing are primarily categorized by the modality of the message, \ie, whether it takes the form of speech, gesture, or written information~\citep{Calvo2010}. In this terminology, affective computing can comprise both unimodal and multimodal analyses. For instance, videos allow for the recognition of facial expressions and vocal tone~\citep{Chen2017, ElAyadi2011, Shan2009}.

The focus of this work is on the unimodal analysis of written materials in English. This choice reflects the prominence of textual materials as a widespread basis for decision-making~\citep{Hogenboom2016}. Illustrative examples are as follows (a detailed review is given later in \Cref{sec:applications}). For instance, the use of affective language as a proxy for emotional closeness can be used to measure the strength of interpersonal ties in social networks~\citep{Marsden2012}. Similarly, marketing utilizes the recognition of emotional states in order to predict the purchase intentions of customers~\citep{Ang2000}, satisfaction with services~\citep{Greaves2013}, and even to measure the overall brand reputation~\citep{Al-Hajjar2015}. In a related context, decision support can leverage affective signals in financial materials in order to suggest trading decisions~\citep{Gilbert2010} or forecast the economic climate~\citep{Nyman2015}. Furthermore, affect can also improve processes and decision-making in the provision of healthcare~\citep{Spiro.2016} or education~\citep{Rodriguez2012}.

Previous research on affective computing has merely utilized methods from traditional machine learning, while recent advances from the field of deep learning -- namely, recurrent neural networks and transfer learning -- have been widely overlooked. However, their use promises further improvements. In fact, techniques from deep learning have become prominent in various decision support activities involving sequential data~\citep[\eg,][]{Evermann2017} 
and especially linguistic materials~\citep[\eg,][]{Kraus2017, Mahmoudi.2018}, where deep learning was able to enhance the performance when deriving decisions from unstructured data. One of the inherent advantages of deep learning is that it can successfully model highly non-linear relationships. 

This work draws upon existing solution techniques from the realm of deep learning \citep{Kraus2017} and applies them to a problem domain different from that of our research objective. First and foremost, we extend existing techniques from the discipline of deep learning to the task of text-based emotion recognition in order to expand the body of knowledge. Following \citet{Kraus2017}, we also utilize long short-term memory networks~(LSTMs) that can make predictions based on running texts of varying lengths. However, affective computing differs substantially from related tasks due to the high number of often imbalanced target labels. Thus, this task requires both customized network architectures and procedures. Hence, its applicability is only made possible through the several methodological innovations that we summarize in the following. 

In order to handle class imbalances in affective computing, we propose the following modifications beyond \citet{Kraus2017}: (i)~bidirectional processing of the text, (ii)~dropout layers as a means of regularization, and (iii)~a weighted loss function. The latter becomes especially critical due to the imbalanced distribution of labels. In fact, without the weighted loss function, the network ends up resembling merely a majority class vote.

We further propose an extension of transfer learning called \emph{sent2affect}. That is, the network is first trained on the basis of sentiment analysis and, after exchanging the output layer, is then tuned to the task of emotion recognition. To the best of our knowledge, this presents a novel strategy for better affective computing as the inductive knowledge transfer is not merely based on a different \emph{dataset}, but a different \emph{task}.

Even though affective computing has gained great traction over the past several years \citep{Ribeiro2016}, there is a scarcity of widely-accepted datasets for text-based emotion recognition that can be used for benchmarking and that facilitate fair comparisons. A relatively small, but more common, dataset was provided by SemEval-2007 and consists of annotated news headlines~\citep{Strapparava2007}. A significantly larger, but underutilized, corpus is composed of affect-labeled literary tales~\citep{Alm2008}. Our literature review notes considerable differences across datasets that vary in their linguistic style, domain, affective dimensions, and the structure of the outcome variable. With regard to the latter, the majority of datasets involve a classification task in which exactly one affective category is assigned to a document, while others request a numerical score across multiple dimensions, \ie, a regression task. Hence, it is a by-product of this research to contribute a holistic comparison that benchmarks different methods across datasets used in prior research. For this purpose, we conducted an extensive search for affect-labeled datasets that serves as the foundation for our computational experiments. As a result, we find that deep learning consistently outperforms the baselines from traditional machine learning. In fact, we observe performance improvements of up to \SI{23.2}{\percent} in F1-score as part of classification tasks and \SI{11.6}{\percent} in mean squared error as part of regression tasks.

The findings of this work have direct implications for management, practice, and research. As such, various application areas of decision support -- such as customer support, marketing, or recommender systems --  can be improved considerably through the use of affective computing. Similarly, all systems with human-computer interactions (\eg chatbots and personal assistants) could further benefit from emotion recognition and a deeper understanding of empathy. In fact, emotion detection could significantly impact and refine all use cases in which sentiment analysis (\ie, only positive/negative polarity) has already proved to be a valuable approach, since these lend themselves to a more fine-grained analysis and decision-making beyond only one dimension. In academia, text-based emotion recognition supports the cognitive and social sciences as a new approach to measuring and interpreting individual and collective emotional states.

The rest of this paper is structured as follows. \Cref{sec:background} reviews earlier works on text-based emotion recognition, including the underlying affect theories, datasets used for benchmarking, and computational approaches. This reveals a research gap with regard to both deep neural networks and transfer learning within the field of affective computing. As a remedy, \Cref{sec:methods} introduces our methods rooted in deep learning, which are then evaluated in \Cref{sec:evaluation}. Based on our findings, we detail implications for both research and management in \Cref{sec:discussion}, while \Cref{sec:conclusion} concludes.

\section{Background} 
\label{sec:background}

We specifically point out that the terms ``sentiment analysis'' and ``affective computing'' are often used interchangeably~\citep{Munezero2014}. However, comprehensive surveys~\citep{Pang2006, Yadollahi2017} recognize clear differences that distinguish each concept: sentiment analysis measures the subjective polarity towards entities in terms of only two dimensions, namely, positivity and negativity. Conversely, affective computing concerns the identification of explicit emotional states and, hence, this approach is also referred to as emotion recognition. The choice of emotional dimensions depends on the underlying affect theory and involves a wide range of mental states such as happiness, anger, sadness, or fear. For reasons of clarity, we strictly distinguish between the aforementioned concepts in our terminology. 

Accordingly, this section first provides an overview of prevalent emotion models as specified by affect theories and, based on their dimensions, reviews computational methods for inferring affective information from natural language. This gives rise to a variety of use cases, which are detailed subsequently. 

\subsection{Affect theory}
\label{sec:affecttheory}

In the field of psychology, there is no consensus regarding a universal classification of emotions~\citep{Frijda1988, Izard2009}, as physiological arousal in the proposed theories varies with causes, cognitive appraisal processes, and context. Yet a conventional approach is to distinguish emotions based on how the underlying constructs are defined. On the one hand, emotions can be defined as a set of discrete states with mutually-exclusive meanings, while, on the other hand, emotions can also be characterized by a combination of numerical dimensions, each associated with a rating of intensity. The categorization into either a discrete set or a combination of intensity labels yields later benefits with regard to computational implementation, as it directly aids in formalizing the different machine learning models.

Categorical emotion models involve a variety of prevalent examples, including the so-called basic emotions. These introduce a discrete set of emotions with innate and universal characteristics~\citep{tomkins1962, Izard1992}. One of the first attempts by \citet{Ekman1987} to classify emotions led to the categorization of six discrete items labeled as basic: namely, anger, disgust, fear, happiness, sadness, and surprise. The model was later extended by \citet{averill1980theories} to include trust and anticipation, resulting in eight basic emotions. An alternative categorization by Tomkins~\citep{tomkins1962,tomkins1963} classifies nine primary affects into positive (enjoyment, interest), neutral (surprise), and negative (anger, disgust, dissmell, distress, fear, shame) expressions.

Dimensional models of emotion locate constructs in a two- or multi-dimensional space~\citep{Poria2017}. Here the assumption of disjunct categories is relaxed such that the magnitude along each dimension can be measured separately~\citep{Russell1980}, yielding continuous intensity scores. Different variants have been proposed, out of which we summarize an illustrative subset in the following. One of the earliest examples is Russell's circumplex model~\citep{Russell1980}, consisting of bivariate classifications into valence and arousal. Depending on the strength of each component, certain regions in the two-dimensional space are given explicit interpretations (such as tense, aroused, excited) according to 28 emotional states. The Wheel of Emotions is an extension of the circumplex model whereby eight primary emotion dimensions are represented as four pairs of opposites: joy versus sadness, anger versus fear, trust versus disgust, and surprise versus anticipation~\citep{Plutchik2001}. Recent approaches introduce complex hybrid emotion models, such as the Hourglass of Emotions~\citep{Cambria2012}, which represents affective states through both discrete categories and four independent, but concomitant, affective dimensions. However, neither the Wheel of Emotions nor the Hourglass of Emotions has yet found its way into common datasets for affective computing.

\subsection{Datasets for benchmarking}
\label{sec:datasets}

\Cref{tab:datatable} provides a holistic overview of datasets used for text-based affective computing. These datasets exhibit fundamentally different characteristics and challenges, as they vary in size, domain, linguistic style and underyling affect theory. We summarize key observations in the following.

In terms of text source, the datasets refer to tasks that utilize narrative materials from classic literature~\citep{Alm2008}, while others are based on traditional media~\citep{Strapparava2007}, and even Twitter or Facebook posts~\citep{Preotiuc-Pietro2016}. Social media, in particular, tends to be informal and subject to variable levels of veracity, especially in comparison with more formal linguistic sources such as newspaper headlines. Similar variations become apparent in terms of where the annotations originate from. For instance, emotion labels can rely upon self-reporting of emotional experiences~\citep{Wallbott.1986} or stem from ex~post labeling efforts via crowdsourcing~\citep{Mohammad2015}.

\begin{table}[t]
	\centering
	\tiny
	\singlespacing
	\makebox[\textwidth]{%
		\renewcommand{\arraystretch}{1.5}
		\begin{tabular}{p{2cm} p{1.5cm} S[table-format=5.0,group-separator={,},group-minimum-digits=3] p{1.7cm} p{2.9cm} S p{2.5cm} p{2.5cm}}
			\toprule
			\textbf{Ref.} & \textbf{Source} & \textbf{Samples} & \multicolumn{4}{c}{\textbf{Emotions}} &  \textbf{Notes} \\
			\cmidrule{4-7}
			& & & \textbf{Annotation} & \textbf{Dimensions} & \textbf{Count} & \textbf{Affect theory} & \\
			\midrule
			\citet{Alm2008} & Literary tales & 1207 & Categorical ($m$-out-of-$n$) & Anger, disgust, fear, happiness, sadness, surprise (pos.), surprise (neg.), neutral & 8 & Basic emotions from \citet{Ekman1987} & Evaluations conventionally draw upon subset where all annotators agree \\
			\citet{Mohammad2015} & Election tweets &  1646 & Categorical ($1$-out-of-$n$) & Anger, anticipation, disgust, fear, joy, sadness, surprise, trust & 8 & Basic emotions from \citet{averill1980theories} &  \\
			\citet{Wallbott.1986} & Self-report of experiences &  7666 & Categorical ($1$-out-of-$n$) & Anger, disgust, fear, guilt, joy, sadness & 7 & Based on basic emotions from \citet{Ekman1987} & Referred to as ISEAR dataset in related literature \\
			\citet{Strapparava2007} & Newspaper headlines  & 1250 & Numerical (for all dimensions)  & Anger, disgust, fear, joy, sadness, surprise; additional valence score & 6 & Basic emotions from \citet{Ekman1987} with valence score according to \citet{Russell1980} & SemEval-2007 (task 14); one numerical score per class \\ 
			\citet{SemEval2018Task1} & General tweets & 7902 & Numerical (single dimension only) & Anger, fear, joy, sadness & 4 & n/a & SemEval-2018 (task 1); for classification tweets with moderate and high emotion \\
			\citet{Preotiuc-Pietro2016} & Facebook posts & 2894 & Numerical & Valence, arousal & 2  & Circumplex model from \citet{Russell1980}  &  \\
			\bottomrule
		\end{tabular}
	}
	\caption{
		Overview of textual datasets used for affective computing in the literature grouped into classification and regression tasks for machine learning.
	}
	\label{tab:datatable}
\end{table}

The majority of datasets were annotated based on categorical emotion models, thereby defining a discrete set of labels. The chosen emotions largely follow suggestions from the different affect theories and predominantly focus on basic emotions (or subsets thereof) due to their prevalence. Even though the number and choice of emotions differ, one can identify four emotions that are especially common as they appear in almost all categorical models: anger, joy (happiness), fear, and sadness.  Some emotions occur more often than others in the usual routines of humans \citep{Plutchik2001,Ekman1987} and one thus obtains datasets ~\citep[\eg,][]{Strapparava2007,Mohammad2015} wherein the relative frequency of emotions is highly unbalanced. This imposes additional computational challenges as classifiers tend to overlook infrequent classes.

In contrast, dimensional models of emotions appear less frequently. Only one dataset, composed of newspaper headlines~\citep{Strapparava2007}, provides a score for each of the six emotion categories. From a methodological point of view, this categorization into dimension-based models requires different prediction models. While categorical models refer to machine learning with single-label classification tasks in the sense that we identify the appropriate item based on a discrete label, dimensional models allow for regression tasks in the sense that we predict a score for every item and emotion.

\subsection{Computational methods}

The automatic recognition of text-based emotions relies upon different computational techniques that comprise lexicon-based methods and machine learning. Due to wealth of approaches, we can only summarize the predominant streams of research in the following and refer to \citet{Calvo2010, Poria2017} for detailed methodological surveys. 

\subsubsection{Lexicon-based methods}

Lexicon-based approaches utilize pre-defined lists of terms that are categorized according to different affect dimensions \citep{Mohammad2012}. On the one hand, these lexicons are often compiled manually, a fact which can later be exploited for keyword matching. For instance, the Harvard~IV dictionary (inside the General Inquirer software) and LIWC provide such lists with classification by domain experts~\citep{Tausczik2010a}. These were not specifically designed for affective computing, but still include psychological dimensions (\eg, pleasure, arousal and emotion in the case of Harvard~IV; anxiety, anger, and sadness for LIWC). The NRC Word-Emotion Association lexicon was derived analogously but with the help of crowdsourcing rather than involving experts from the field of psychology research~\citep{Mohammad2013}. The latter dictionary includes 10 granular categories such as anticipation, trust, and anger. 

In order to overcome the need for manual dictionary creation, heuristics have been proposed to construct affect-related wordlists. Common examples include the WordNet-Affect dictionary, which starts with a set of seed words labeled as affect and then assigns scores to all other words based on their proximity to the seed words \citep{Strapparava2004a}. However, the resulting affect dictionary includes only general categories of mood- or emotion-related words, rather than further distinguishing the type of emotion. More recent methods operate, for instance, via mixture models~\citep{Bandhakavi2017}, fuzzy clustering \citep{Poria2014}, or by incorporating word embeddings~\citep{Li2017}. The precision of dictionaries can further be improved by embedding these in linguistic rules that adjust for the surrounding context.

Dictionary-based approaches are generally known for their straightforward use and out-of-the-box functionality. However, manual labeling is error-prone, costly, and inflexible as it impedes domain customization. Conversely, the vocabulary from the heuristics is limited to a narrow set of dimensions that were selected a priori and, as a result, this procedure has difficulties when generalizing to other emotions~\citep[\cf][]{Agrawal2012}. 

\subsubsection{Machine learning}

Machine learning can infer decision rules for recognizing emotions based on a corpus of training samples with explicit labels~\citep{Danisman2008, Chaffar2011}. This can overcome the aforementioned limitations of lexicon-based methods concerning scalability and domain customization. Moreover, it can also learn implicit signals of emotions, since findings from a comprehensive, comparative study suggest that affect is rarely communicated through emotionally-charged lexical cues but rather via implicit expressions~\citep{Balahur2012a}. 

Previous research has experimented with different models for inferring affect from narrative materials. Examples include methods that explicitly exploit the flexibility of machine learning, such as random forests~\citep[\eg,][]{Gordeev2016} and support vector machines~\citep[\eg,][]{Chatzakou2017,Danisman2008}, both of which have commonly been deployed in literature. Studies have shown that random forests tends to compute faster, while support vector machines yield superior performance~\citep{Chatzakou2017}. These classifiers are occasionally, but infrequently, restricted to the subset of affect cues from emotion lexicons~\citep{Bandhakavi2017}. However, the more common approach relies upon general linguistic features, \ie, bag-of-words with subsequent tf-idf weighting~\citep{Alm2005,Strapparava2007}. Consistent with these works, we later draw upon machine learning models (\ie, random forest and support vector machine) together with tf-idf features as our baseline.

\subsubsection{Deep learning}

In the following, we discuss the few attempts at applying deep learning to affective computing, but find that actual performance evaluations are scarce. The approach in \citet{Gordeev2016} predicts aggression expressed through natural language using convolutional neural networks with a sliding window and subsequent max-pooling. However, this approach is subject to several limitations as the network is designed to handle only a single dimension (\ie, aggression) and it is thus unclear how it generalizes across multi-class predictions or even regression tasks that appear in dimensional emotion models. Even though the approach utilizes a \textquote{deep} network, its network architecture can only handle texts of predefined size, analogous to traditional machine learning. In this respect, it differs from recurrent networks, which iterate over sequences and thus can handle texts of arbitrary size.

The work in~\citet{Felbo2017a} utilizes an LSTM that is pretrained with tweets based on the appearance of emoticons; however, this work does not report a comparison of their LSTM against a baseline from traditional machine learning. A different approach \citep{Gupta2017} utilizes a custom LSTM architecture in order to assign emotion labels to complete conversations in social media. However, this approach is tailored to the specific characteristics and emotions of this type of conversational-style data. In addition, the conclusion from their numerical experiments cannot be generalized to affective computing, since the authors labeled their dataset through a heuristic procedure and then reconstructed this heuristic with their classifier. Closest to our approach are experiments that include an LSTM for intensity estimation of emotions \citep{Goel.2017,Lakomkin.2017,Meisheri.2017,Zhang.2017}, but the results are limited to regression tasks where the presence of specific affective dimensions is given a priori.  

Up to this point, the potential performance gains from using recurrent neural networks as the state of the art in deep learning have not yet been studied in relation to text-based emotion recognition. This fact was also noted in a recent literature survey~\citep{Poria2017}. 

\subsection{Transfer learning}

Transfer learning is a technique whereby knowledge from a source domain is leveraged in order to improve performance in a (possibly different) target domain. It is often used to overcome the constraints of limited training data, as well as for tasks that are sensitive to overfitting~\citep{Pan.2010}. A straightforward approach to transferring knowledge in natural language applications is to draw upon pretrained word embeddings~\citep{Kraus2017}. This approach merely requires an additional dataset without labels as it operates in unsupervised fashion. However, it only facilitates the representation of words and fails to help learning parameters inside the neural network. 

More complex strategies can even utilize labels and perform transfer learning from a source to a target dataset. The underlying transfer can occur either concurrently or sequentially:
\begin{itemize}
\item The former trains two networks concurrently on both the source and the target task with shared parameters. For instance, one network learns to translate sentences, while the other recognizes named entities \citep{Mou.2016}. This is known to help the network concentrate on a shared understanding and, in practice, puts emphasis on more abstract relationships.
\item The latter sequential procedure first trains a network on a source dataset and, in a second step, applies the network to the target dataset in order to fine-tune the network parameters \citep[\eg,][]{Kratzwald.2018}. This is often accompanied by minor modifications to network architectures (\eg, by replacing the prediction layer). While such an approach seems intriguing, it is impeded by the heterogeneous nature of baseline datasets for emotion recognition. 
\end{itemize}
However, natural language applications often lack suitable source datasets \citep{Mou.2016}. As a remedy, we propose sent2affect: that is, we employ not only a different \emph{dataset} but also a different \emph{task} (namely, sentiment analysis). To the best of our knowledge, this presents the first work on affective computing that attempts to accomplish an inductive knowledge transfer across tasks.

\section{Methods} 
\label{sec:methods}

% overview

\begin{figure}[!ht]
\small
\includegraphics[width=1\textwidth]{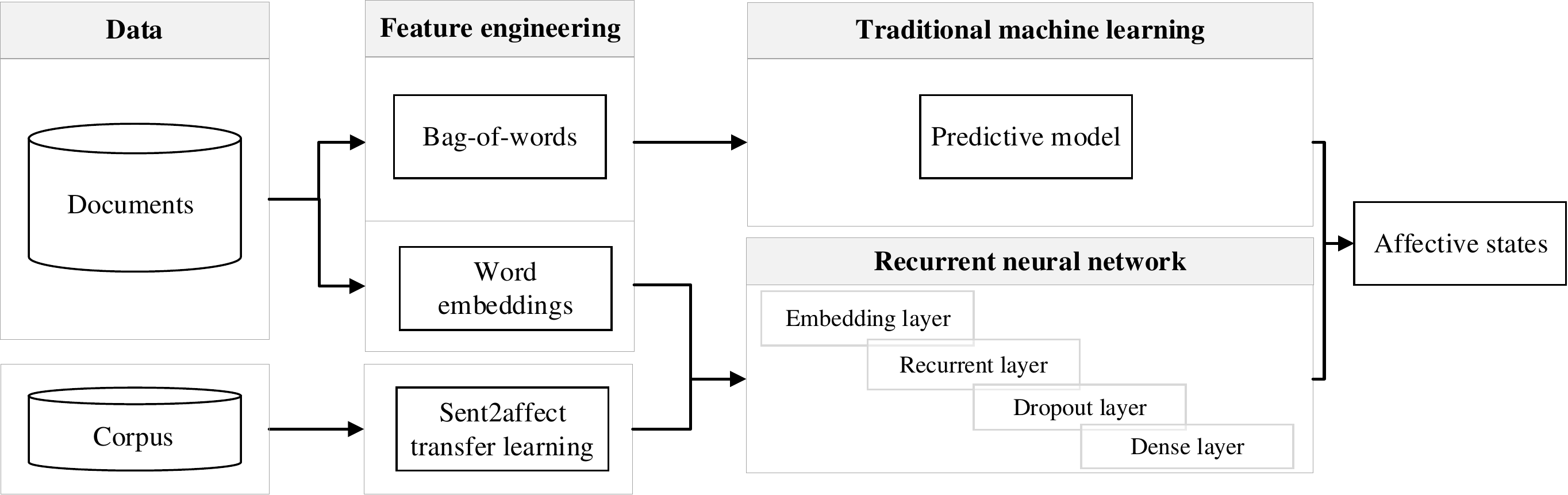}
\caption{Illustrative pipeline for inferring affective states from narrative materials. This can either happen through (i)~traditional machine learning with feature engineering or, as proposed in this work, (ii)~deep recurrent neural networks, optionally in conjunction with our proposed sent2affect transfer learning.}
\label{fig:pipeline}
\end{figure}

This section presents our methods for inferring emotional states from narrative contents. We first summarize our baselines from traditional machine learning and deep learning, while the inherent nature of affective computing requires us to come up with multiple innovations concerning the network architecture. Our proposed advances are detailed in \Cref{sec:new_deep_learning}. Finally, we detail our novel approach to transfer learning, called sent2affect, whereby knowledge from the related task of sentiment analysis is applied to emotion recognition. \Cref{fig:pipeline} illustrates this pipeline.

\subsection{Benchmarks}

\subsection{Baselines from traditional machine learning}

Traditional machine learning can only learn from a fixed-size vector of features and, for this purpose, features for machine learning are commonly built upon bag-of-words. The frequencies are further weighted by the tf-idf scheme in order to measure the relative importance of terms to a document within a corpus. Mathematically, the measure of term importance is obtained by computing the product of the term frequency and the inverse document frequency. This approach serves as a widely-accepted benchmark against which algorithms for natural language processing are evaluated.

The aforementioned features are then fed into the actual predictive models from traditional machine learning. Here we chose two approaches for both classification and regression as our baseline models: namely, random forest and support vector machine (\ie, a support vector regression for predictive numerical scores). These are known for their superior performance in previous studies~\citep[\eg,][]{Chatzakou2017}. Moreover, both approaches entail high flexibility when modeling non-linear relationships and demonstrate high accuracy even in settings where the number of potential features exceeds the number of observations. 

\subsubsection{Baselines from na{\"i}ve deep learning}

Deep learning has triggered a paradigm shift in machine learning \citep{Kraus.2018} since it has yielded unprecedented performance results, especially for natural language processing. The theoretical argument for this is that recurrent neural networks from deep learning can iterate over the individual words of a sequence with arbitrary length. Here the input directly consists of words $x_1, \ldots, x_N$ and thus circumvents the need for feature engineering (\eg, creating bag-of-words with tf-idf) as used in traditional machine learning. As a result, recurrent neural networks store a lower-dimensional representation of the input sequence that encodes the whole document and can even maintain the actual word order with long-ranging semantics \citep{Kraus.2018}. For this reason, recurrent neural networks differ from traditional machine learning, which can only adapt to short texts due to the use of $n$-grams. 

We draw upon \citet{Kraus2017} as the basis for our deep neural network architecture. This basic model consists of three layers: (a)~an embedding layer that maps words in one-hot encoding onto low-dimensional vectors, (b)~a recurrent layer to pass information on between words, and (c)~a final dense layer for making the actual prediction. All three layers are described in detail in the online appendix. We experimented with this approach, but found that its performance is almost identical to a majority class vote. Therefore, we refrain from reporting the exact results; instead, we focus on the following improvements. 

\subsection{Proposed deep neural networks for affective computing}
\label{sec:new_deep_learning}

Using the aforementioned deep learning architectures is non-trivial for the following reasons. First, they are not suited to the small datasets from affective computing and typically lead to severe overfitting. Hence, we propose the use of a dropout layer as a form of regularization. Second, our task involves complex, open-domain language, which benefits further from bidirectional processing. Third, severe class imbalances are addressed by a weighted loss function. This loss function treats each class equally in order to avoid biases towards certain classes. Altogether, these extensions were necessary for using deep learning in our research setting.

\subsubsection{Dropout layer}

Deep neural networks can easily consist of up to millions of free parameters and, consequently, these models run the risk of overfitting. This is especially a problem when the training data is scarce. As a remedy, the weights in the network are regularized by randomly dropping out a certain share of neurons in order to improve the generalizability of the network. This prevents the neurons from co-adapting too much during training \citep{Srivastava.2014}. We use dropout within the recurrent layer; that is, we randomly drop out connections between the recurrent LSTM cells. Dropout is disabled, \ie, all neurons are used, during test time in order to leverage the full predictive power of the learned parameters (cf. the online appendix for a detailed specification). Furthermore, we apply dropout between the output of the recurrent layer and the input to the prediction layer. 

\subsubsection{Bidirectional processing}

To further improve the predictive performance of the base model, we draw upon so-called bidirectional recurrent layers, which have shown success in various other domains. That is, we use not only one but two LSTM layers to read the text. While one layer processes the text from left to right, a second one processes the text from right to left. More formally, let $h_1$ determine the hidden state of the LSTM network that processes the input in the forward direction and $h_2$ the hidden state of the LSTM that reads the text backwards. We then use the concatenation of both hidden states, \eg, $[h_1, h_2]$, as input for the final prediction layer. Thus we are able to cover long- and short-term dependencies in both directions. We later abbreviate the bidirectional LSTM via BiLSTM and additionally run separate experiments for comparing the performance across the LSTM and BiLSTM.

\subsubsection{Weighted loss functions for unbalanced data}

Affective computing commonly involves multiple, highly imbalanced target labels. Using a na{\"i}ve loss function in this case would optimize towards the majority class and thus result in a performance similar to a majority vote. Such problems are typically addressed by over- or undersampling, yet these approaches yielded only marginal improvements in our experiments. As an alternative, we suggest the use of a weighted loss function. This multiplies the error of each data point with a weight that is the inverse size of the corresponding class. 

Assume a training sample $x_i$ with ground-truth label $y_i$, and $p_{ik}$ denoting the output of the prediction layer, \eg, the probability of $x_i$ belonging to class $k$. Then the weighted loss for $x_i$ is calculated via 
\begin{equation}
\mathcal{L}(x_i, \theta) = w_{i}  \sum_{k=1}^{K} \mathds{1}_{y_i=k} \log p_{ik} 
\end{equation}
with $\mathds{1}$ denoting the indicator function. The weight $w_{i}$ for input $x_i$ depends solely on its ground truth label $y_i$ and, similar to \citet{King.2001}, is calculated as
\begin{equation}
	w_{i} = \frac{N}{K \sum_{j} \mathds{1}_{y_j=y_i}} ,
\end{equation}
where $K$ denotes the total number of classes and $N$ the number of samples.

\subsection{Sent2affect approach to transfer learning across tasks}

Due to the large number of degrees-of-freedom, training deep neural networks is often associated with challenges (\eg, overfitting, ineffective generalization). In practice, this is encountered by large datasets in order to prevent overfitting and, hence, a different strategy is often applied when handling smaller datasets such as those in our experiments. Here the idea is to implement transfer learning, \ie, the inductive transfer of knowledge from a different, yet related, task to the problem under investigation. In our case, we develop a novel approach, sent2affect, as detailed in the following.

The choice of the source task is non-trivial and it is mainly tasks of a semantically similar nature that result in the transferability of the network. For this purpose, we suggest the use of sentiment analysis as a related task, since it shares a certain similarity in the sense that positive and negative polarity is inferred from linguistic materials; however, sentiment analysis differs from affective computing, as it does not address affective dimensions or emotional states. The relatedness between both tasks enables the network to infer similar representation for both.

Formally, our approach to transfer learning optimizes the weights of a neural network for a target task $\mathcal{T}$ and dataset $\mathcal{D_T}$ based on a different, yet related, source task $\mathcal{S}$ with dataset $\mathcal{D_S}$. After optimizing the parameters of our network for $\mathcal{S}$ on $\mathcal{D_S}$ we replace the task-specific prediction layer of the network to yield predictions for our target task $\mathcal{T}$. Therefore, we utilize the estimated parameters as an initial value for further optimization with the help of the actual dataset $\mathcal{D_T}$~\citep{Pan.2010}. The pseudocode of the overall process is stated in \Cref{alg:transfer}.

In our experiments, we utilize a large-scale, public dataset\footnote{Kaggle: Twitter sentiment analysis, retrieved from \url{https://www.kaggle.com/c/twitter-sentiment-analysis2}, March~21, 2018.} as a basis for knowledge induction. This dataset finds widespread application in sentiment analysis and includes about \num[group-separator={,},group-minimum-digits=3]{100000} samples labeled according to positive or negative sentiment. We then optimize the deep neural network with the goal of predicting the underlying sentiment scores. The resulting coefficients of the network are further trained with an actual dataset from affective computing. Here the differences in the data type of the prediction outcome (\ie, computing a positivity/negativity score versus affective dimensions) are handled by removing the dense layer and, instead, amending a new prediction layer that targets the new output. As a result, the majority of weights benefits from transfer learning, while only the neurons in the prediction layer are training after a random initialization. The intuition of this approach is as follows: deep neural networks generally contain multiple layers, where layers closer to the final prediction layers are supposed to encode the original input at a higher level of abstraction. 

\begin{algorithm}[H]
	\caption{sent2affect transfer learning}
	\label{alg:transfer}
	\footnotesize
	\renewcommand{\algorithmicrequire}{\textbf{Input:}}
	\renewcommand{\algorithmicensure}{\textbf{Output:}}
	\begin{algorithmic}[1]
		\Require Given training data $\mathcal{D_T}$ for the affective computing task $\mathcal{T}$ and additional corpus $\mathcal{D_S}$ for sentiment analysis $\mathcal{S}$
		\State $m \gets $ Initialize recurrent neural network (\ie consisting of recurrent layer $f$, dense layer $\psi$, \ldots)
		\State $m \gets $ Estimate parameters w.r.t. $\mathcal{S}$ using $\mathcal{D_S}$
		\State $\psi \gets $ Replace dense layer with randomly-initialized dense layer according to the dimensions of $\mathcal{T}$
		\State $\psi \gets $ Fine-tune $\psi$ w.r.t. $\mathcal{T}$ using $\mathcal{D_T}$
		\State \textbf{return} Recurrent neural network $m$
	\end{algorithmic}
\end{algorithm}

\subsection{Model estimation}

Consistent with previous research~\citep{manning1999foundations}, we tokenize each document, convert all characters to lower-case, and remove punctuation, numbers, and stop words. Moreover, we perform stemming, which maps inflected words onto a base form; \eg, \emph{\textquote{played}} and \emph{\textquote{playing}} are both mapped onto \emph{\textquote{play}}. We conducted all pre-processing operations to yield bag-of-words representations by using the natural language tookit NLTK. 

For those datasets with no designated test set, we introduced a random $80/20$ split in training and test data. For the random forest classifier, we manually optimized over the number of trees, maximum number of features for every split, and the depth. For the support vector classifier, we conducted an extensive grid-search over the hyperparameters following \citet{hsu2003practical}. In detail, we experimented with linear, radial basis function, and sigmoid kernels, optimizing the cost $C$ over $2^{-5}, 2^{-3},\ldots,2^{15}$ and the radius parameter $\gamma$ over $2^{-15},2^{-13},\ldots,2^3$. For unbalanced datasets, we weighted the loss function by class frequency in order to prevent models from predicting the majority classes only. 

We used different deep learning models. Depending on the specification, we used pre-trained GloVe\footnote{The pre-trained word embeddings can be retrieved from \url{http://nlp.stanford.edu/data/glove.6B.zip}.} embeddings or randomly-initialized embeddings (which are learned jointly during the training phase). The models were trained using the Adam optimizer, whereby the process was stopped once we noted an increase in the validation error. For reasons of reproducibility, we report the performance metrics averaged over 10 independent runs.

\section{Evaluation} 
\label{sec:evaluation}

This section reports our computational experiments evaluating the improvements gained by using deep neural networks (and especially transfer learning) for affective computing. Here we draw upon all datasets from \Cref{tab:datatable} and, according to the type of the underlying affect theory, we divide the performance measurements into classification and regression tasks. 

\subsection{Classification according to categorical emotion models}

% what we did: experiment overview
We begin with classification tasks according to categorical emotion models, where the objective is to predict the predominant emotion(s). We follow previous literature \citep[\eg,][]{Chatzakou2017,Danisman2008} and analogously choose two baselines prevalent in traditional machine learning: namely, the random forest classifier and the support vector machine. Both are fed with bag-of-words with tf-idf weighting, whereas the proposed deep neural networks circumvent the need for feature engineering. Here we compare variants that extend the LSTM\footnote{We use the acronym LSTM when referring to the unidirectional model. Whenever we refer to the bidirectional LSTM model, we use the explicit designation BiLSTM.} with bidirectional encodings and pretrained word embeddings. The resulting performance is listed in \Cref{tab:results_classification}, where we account for unbalanced distributions of labels by using the weight-averaged F1-score. The F1-score for a single class is given by the harmonic mean of precision and recall, \ie,
\begin{equation}
\text{F1} = 2 \frac{\text{precision} \cdot \text{recall}}{\text{precision} + \text{recall}} .
\end{equation}
In addition, we report sensitivity and specificity scores. The sensitivity of a single class equals the recall, while the specificity measures the fraction of true negatives. Similar to the F1-score, we calculate both independently for each class, \ie,  
\begin{equation}
\text{sensitivity} = \mathit{TP}/(\mathit{TP}+\mathit{FN}) 
\qquad \text{and} \qquad
\text{specificity} = \mathit{TN}/(\mathit{TN}+\mathit{FP}),
\end{equation}
where the number of true positives and true negatives is denoted by $\mathit{TP}$ and $\mathit{TN}$, and the number of false positives and false negatives is denoted by $\mathit{FP}$ and $\mathit{FN}$. For the final scores, we average over all classes weighted by the class size.

% results 

Our results in \Cref{tab:results_classification} consistently reveal superior performance through the use of deep learning. We observe that, regardless of the architecture, models with pre-trained GloVe embeddings outperform their counterparts with randomly-initialized word embeddings. In fact, the use of pre-trained word embeddings yields performance improvements over the best baseline in $9$ out of $10$ experiments. An explanation stems from the fact that embeddings which have not been pre-trained result in considerably more degrees-of-freedom and thus a greater chance of overfitting. Our initial expectations are met as the imposed dropout layers and loss-weighting successfully diminish the problem of overfitting. Furthermore, our imposed architectural enhancements surpass the performance of previous deep learning architectures, such as that proposed by \cite{Kraus2017}. As such, the bidirectional recurrent layers outperform the variant with a unidirectional layer in four out of five experiments, yielding the only architecture that consistently outperforms the traditional baseline on all datasets, with improvements between \SI{1.6}{\percent} and \SI{23.2}{\percent} across the datasets. We experimented with the na\"{i}ve network from \citet{Kraus2017}, but it failed in three out of five datasets resulting in merely predicting the majority class; hence, we omitted the results.

The performance gains from our proposed architectural improvements are a result of the class imbalance and the language noise of the source. For instance, the highest relative improvement over traditional machine learning is achieved in the case of the dataset of headlines~\citep{Strapparava2007}, constructed of four equally-sized classes and proper English. On the other hand, the dataset of election tweets \citep{Mohammad2015}, which is composed of highly unbalanced classes and considerable language noise, yields the lowest improvement.

\Cref{tab:sensitivity_specificity} reports sensitivity and specificity scores as an additional robustness check. The results confirm our findings, \ie, we witness the largest performance improvements for datasets with less noise. For the election tweet dataset \citep{Mohammad2015}, the best bidirectional LSTM model achieves a sensitivity of $56.9$, while the best baseline achieves a slightly better score of $57.1$. We can significantly strengthen our results for this challenging dataset by applying transfer learning, as reported in \cref{sec:results_transfer_learning}.

% table with all results

\begin{table}[h]
\centering
\scriptsize
\sisetup{input-symbols={()\,\%}}
\makebox[\textwidth]{
\begin{tabular}{l cc @{\hskip 0.3cm} cc @{\hskip 0.3cm} cc}
\toprule
\textbf{Dataset} & \multicolumn{2}{c}{\textbf{\mcellt{Baseline: traditional\\ machine learning}}} & \multicolumn{2}{c}{\textbf{Deep learning}} & \multicolumn{2}{c}{\textbf{\mcellt{Pre-trained\\ word embeddings}}} \\
\cmidrule(lr){2-3}\cmidrule(lr){4-5}\cmidrule(lr){6-7}
& \textbf{Random forest} & \textbf{SVM} & \textbf{LSTM} & \textbf{BiLSTM} &  \textbf{LSTM} & \textbf{BiLSTM} \\
\midrule

{Literary tales} \citep{Alm2008}  &    63.2   &   64.7   &  63.0   &  61.6 &   \bfseries 67.9  &      \bfseries 68.8          \\
& & & \multicolumn{1}{c}{(\SI{-2.6}{\percent})} & \multicolumn{1}{c}{(\SI{-4.8}{\percent})} & \multicolumn{1}{c}{(\SI{+4.9}{\percent})} & \multicolumn{1}{c}{(\SI{+6.3}{\percent})} \\

{Election tweets} \citep{Mohammad2015} & 55.0  & 56.8  & 54.5  & 54.8   & 55.8     & \bfseries 57.7       \\
& & & \multicolumn{1}{c}{(\SI{-4.0}{\percent})} & \multicolumn{1}{c}{(\SI{-3.5}{\percent})} & \multicolumn{1}{c}{(\SI{-1.8}{\percent})} & \multicolumn{1}{c}{(\SI{+1.6}{\percent})} \\

\lcellt{{ISEAR} \citep{Wallbott.1986}, \ie, self-reported}     & 47.0 & 54.3 & 54.2 &\bfseries 55.8 & \bfseries 57.7 & \bfseries 56.9   \\
\quad experiences & & & \multicolumn{1}{c}{(\SI{-0.2}{\percent})} & \multicolumn{1}{c}{(\SI{+1.5}{\percent})} & \multicolumn{1}{c}{(\SI{+6.3}{\percent})} & \multicolumn{1}{c}{(\SI{+4.8}{\percent})} \\

{{Headlines}} \citep{Strapparava2007}        & 35.8  & 35.3  & \bfseries 39.2 & \bfseries39.8   &\bfseries 41.7          & \bfseries 44.1         \\
& & & \multicolumn{1}{c}{(\SI{+9.5}{\percent})} & \multicolumn{1}{c}{(\SI{+11.2}{\percent})} & \multicolumn{1}{c}{(\SI{+16.5}{\percent})} & \multicolumn{1}{c}{(\SI{+23.2}{\percent})} \\ 

{General tweets} \citep{SemEval2018Task1} & 52.6          & 54.2        &\bfseries 56.0      & \bfseries55.5    & \bfseries 57.7     &\bfseries58.2        \\
& & & \multicolumn{1}{c}{(\SI{+3.3}{\percent})} & \multicolumn{1}{c}{(\SI{+2.4}{\percent})} & \multicolumn{1}{c}{(\SI{+6.5}{\percent})} & \multicolumn{1}{c}{(\SI{+7.4}{\percent})} \\

\bottomrule
\end{tabular}
}
\caption{Holistic comparison of traditional machine learning and recurrent neural networks (with optional GloVe word embeddings) for affective computing, that is, models as classification tasks. Here the outcome variable represents a single label according to predefined categorical emotion model. Accordingly, the performance is measured based on the F1-score; \ie, the higher the better. All models that outperform the best baseline model are highlighted in bold. The percentage changes refer to the relative improvement over the best baseline from traditional machine learning.}
\label{tab:results_classification}
\end{table}

\begin{table}[htb]
	\centering
	\scriptsize
	\sisetup{input-symbols={()\,\%}}
	\makebox[\textwidth]{
		\begin{tabular}{l cc @{\hskip 0.3cm} cc @{\hskip 0.3cm} cc}
			\toprule
			\textbf{Dataset} & \multicolumn{2}{c}{\textbf{\mcellt{Baseline: traditional\\ machine learning}}} & \multicolumn{2}{c}{\textbf{Deep learning}} & \multicolumn{2}{c}{\textbf{\mcellt{Pre-trained\\ word embeddings}}} \\
			\cmidrule(lr){2-3}\cmidrule(lr){4-5}\cmidrule(lr){6-7}
			& \textbf{Random forest} & \textbf{SVM} & \textbf{LSTM} & \textbf{BiLSTM} &  \textbf{LSTM} & \textbf{BiLSTM} \\
			\cmidrule(lr){2-2}\cmidrule(lr){3-3}\cmidrule(lr){4-4}\cmidrule(lr){5-5}\cmidrule(lr){6-6}\cmidrule(lr){7-7} 
			& Sens. \quad Spec. & Sens. \quad Spec. & Sens. \quad Spec. & Sens. \quad Spec. & Sens. \quad Spec. & Sens. \quad Spec. \\ 
			\midrule
			
			{Literary tales} \citep{Alm2008}                 &    64.0\quad87.2                   &     66.1\quad87.4       &       63.1\quad88.8     &    62.0\quad88.8            &      \textbf{68.4}\quad90.6                &       68.2\quad\textbf{91.4}                   \\
			{Election tweets} \citep{Mohammad2015}                 & 53.8\quad79.2                   & \textbf{57.1}\quad78.7         & 52.3\quad84.7          & 52.3\quad\textbf{84.9}            & 54.0\quad83.5       &  56.9\quad82.9       \\
			\lcellt{{ISEAR} \citep{Wallbott.1986}, }                          & 45.0\quad90.3                 & 53.7\quad92.4         & 54.2\quad92.3          & 56.0\quad92.6            & \textbf{57.6}\quad\textbf{93.0}        & 57.0\quad92.8                 \\
			\quad \ie, self-reported experiences & & & & & & \\
			{Headlines} \citep{Strapparava2007}        & 35.6\quad86.2                   & 35.0\quad84.2         & 39.8\quad83.8          & 40.3\quad84.1         & 40.0\quad\textbf{86.6}                 & \textbf{44.3}\quad85.6           \\
			{General tweets} \citep{SemEval2018Task1} & 52.4\quad84.1                   & 53.9\quad84.6         & 55.6\quad85.2          & 55.1\quad85.0            & 57.3\quad85.8         & \textbf{57.9}\quad\textbf{85.9}             \\
			
			\bottomrule
		\end{tabular}
	}
	\caption{Additional comparison of sensitivity and specificity. The highest values for both are highlighted for each dataset.}
	\label{tab:sensitivity_specificity}
\end{table}

\subsection{Regression according to dimensional affect models}

Depending on the affect theory, one can also model emotional categories according to dimensional ratings and, as a result, this is implemented as a regression task, where the intensity of emotional states is predicted. We choose the same baselines as in the previous experiments and compare them to deep neural networks. All models are evaluated based on the mean squared error~(MSE).

\Cref{tab:results_regression} reports our results. These show a consistent improvement of up to \SI{11.6}{\percent} as a result of using deep learning as compared to traditional machine learning. Similar to the classification task, our findings identify the BiLSTM with pre-trained word embeddings as the superior method in all seven experiments. We further note that the BiLSTM appears to outperform the unidirectional LSTM in all experiments. The relative performance increases vary between the different affective dimensions.

\begin{table}[htb]
\centering
\scriptsize
\makebox[\textwidth]{
\begin{tabular}{p{3cm} l SS @{\hskip 0.3cm} SS @{\hskip 0.3cm} SS}
\toprule
\textbf{Dataset} & \textbf{Scale} & \multicolumn{2}{c}{\textbf{\mcellt{Baseline: Traditional\\ machine learning}}} & \multicolumn{2}{c}{\textbf{Deep learning}} & \multicolumn{2}{c}{\textbf{\mcellt{Pre-trained\\ word embeddings}}} \\
\cmidrule(lr){3-4}\cmidrule(lr){5-6}\cmidrule(lr){7-8}
                           && \textbf{Random forest} & \textbf{SVM} & \textbf{LSTM}     & \textbf{BiLSTM}  &  \textbf{LSTM}   & \textbf{BiLSTM}    \\   \midrule
Headlines \citep{Strapparava2007}  \\
\quad \emph{Valence}  & $-100$\ldots$100$      & 1906.0           & 1927.3       & \bfseries 1870.9  & \bfseries 1896.3 & \bfseries 1792.7 & \bfseries  1791.2  \\  
 && & & \multicolumn{1}{c}{(\SI{-1.8}{\percent})}
 & \multicolumn{1}{c}{(\SI{-0.5}{\percent})}  & \multicolumn{1}{c}{(\SI{-5.9}{\percent})}  & \multicolumn{1}{c}{(\SI{-6.0}{\percent})}            \\   \midrule
Facebook posts \citep{Preotiuc-Pietro2016} \\
\quad \emph{Valence}  & 0\ldots10      & 1.030            & 0.951        & 1.007             & 0.990            & \bfseries 0.911  & \bfseries  0.901   \\
 && &   & \multicolumn{1}{c}{(\SI{+5.9}{\percent})}
 & \multicolumn{1}{c}{(\SI{+4.1}{\percent})}  & \multicolumn{1}{c}{(\SI{-4.2}{\percent})}  & \multicolumn{1}{c}{(\SI{-5.2}{\percent})}            \\
\quad \emph{Arousal}   & 0\ldots10     & 3.960            & 3.616        & \bfseries 3.519   & \bfseries 3.550  & \bfseries 3.379  & \bfseries  3.346   \\
 && &   & \multicolumn{1}{c}{(\SI{-2.7}{\percent})}
 & \multicolumn{1}{c}{(\SI{-1.8}{\percent})}  & \multicolumn{1}{c}{(\SI{-6.6}{\percent})}  & \multicolumn{1}{c}{(\SI{-7.5}{\percent})}            \\   \midrule
General tweets \citep{SemEval2018Task1}    \\ 
\quad \emph{Anger}   & 0\ldots1       & 0.0314           & 0.0323       & 0.0330            & 0.0330         & \bfseries 0.0284 & \bfseries  0.0281  \\
 && &   & \multicolumn{1}{c}{(\SI{+5.1}{\percent})}
 & \multicolumn{1}{c}{(\SI{+5.1}{\percent})}  & \multicolumn{1}{c}{(\SI{-9.5}{\percent})}  & \multicolumn{1}{c}{(\SI{-10.5}{\percent})}           \\
\quad \emph{Fear}    & 0\ldots1    & 0.0245           & 0.0226       & 0.0238            & 0.0230           & \bfseries 0.0224 & \bfseries 0.0222   \\
 && & & \multicolumn{1}{c}{(\SI{+5.3}{\percent})}
 & \multicolumn{1}{c}{(\SI{+1.8}{\percent})}  & \multicolumn{1}{c}{(\SI{-0.9}{\percent})}  & \multicolumn{1}{c}{(\SI{-1.8}{\percent})}            \\
\quad \emph{Joy}    &  0\ldots1       & 0.0339           & 0.0294       & \bfseries 0.0277  & \bfseries 0.0275 & \bfseries 0.0262 & \bfseries  0.0260  \\
 && & & \multicolumn{1}{c}{(\SI{-5.8}{\percent})}
 & \multicolumn{1}{c}{(\SI{-6.5}{\percent})}  & \multicolumn{1}{c}{(\SI{-10.9}{\percent})} & \multicolumn{1}{c}{(\SI{-11.6}{\percent})}         \\
\quad \emph{Sadness}  & 0\ldots1     & 0.0294           & 0.0274       & 0.0281 & \bfseries 0.0268 & \bfseries 0.0246 & \bfseries  0.0243 \\ 
 && & & \multicolumn{1}{c}{(\SI{+2.5}{\percent})}
 & \multicolumn{1}{c}{(\SI{-2.1}{\percent})}  & \multicolumn{1}{c}{(\SI{-10.2}{\percent})} & \multicolumn{1}{c}{(\SI{-11.3}{\percent})}         \\ \bottomrule
\end{tabular}
}

\caption{Holistic comparison of traditional machine learning and recurrent neural networks (with optional GloVe word embeddings) for affective computing, that is, models as regression tasks. Here the outcome variable represents the intensity according to predefined affective dimensions. Accordingly, the performance is measured based on the mean squared error (MSE); \ie, the lower the better. The best-performing model for each dataset is highlighted in bold. The percentage changes refer to the relative improvement over the best baseline from traditional machine learning. We point out that the first task exhibits higher errors due to the different scale of the outcome variable.}
\label{tab:results_regression}
\end{table}

\subsection{Transfer learning via sent2affect}
\label{sec:results_transfer_learning}

The previous experiments revealed consistent improvements through the use of deep learning; however, several benchmark datasets entail only a fairly small set of samples, which could impede the training of deep neural networks. For instance, the dataset of inferring emotions from election tweets~\citep{Mohammad2015} comprises only 1,646 samples for training. A potential remedy is utilizing large-scale datasets from other tasks and then inducing knowledge to affective computing. More precisely, we now experiment with the potential performance improvements to be gained by additionally applying our transfer learning approach \textquote{sent2affect}. By inducing network parameters from sentiment analysis to affective computing, we benefit from the considerably larger datasets that are used in sentiment analysis, since the sentiment dataset consists of about \num[group-separator={,},group-minimum-digits=3]{100000} tweets that are associated with positive and negative labels. 

\Cref{tab:results_transfer_learning} compares our transfer learning approach against two baselines: (i)~a na{\"i}ve BiLSTM and (ii)~the transfer learning approach of \citet{Kraus2017}, where only GloVe word-embeddings are pre-trained. We choose the election tweets~\citep{Mohammad2015} and general tweets~\citep{SemEval2018Task1} datasets to demonstrate how we can transfer the knowledge from thousands of sentiment-labeled tweets to the task of emotion recognition. Furthermore, na{\"i}ve deep learning alone yields an inferior performance. While the BiLSTM with pre-trained word embeddings has previously represented the best-performing architecture, we still observe that transfer learning yields additional improvements. These amount to \SI{6.6}{\percent} for the election tweets and \SI{5.6}{\percent} for the general tweets. Evidently, transfer learning can successfully benefit from the large-scale dataset for sentiment analysis and, as a result, optimizes the neuron weights such that these find a more generalizable representation of emotion-laden materials.

\begin{table}[h]
\centering
\scriptsize
{
\begin{tabular}{p{3cm} SSS}
\toprule
\textbf{Dataset}   & \textbf{\mcellt{Na{\"i}ve\\ BiLSTM}} & \textbf{\mcellt{BiLSTM\\ (pre-trained embeddings)}} & \textbf{\mcellt{Transfer\\ learning\\  sent2affect}} \\
\midrule
Election tweets \citep{Mohammad2015}    & 54.8   & 57.7                                      & \bfseries 58.4	                         \\
&    & \multicolumn{1}{c}{(\SI{+5.3}{\percent})} & \multicolumn{1}{c}{(\SI{+6.6}{\percent})}  \\
General tweets \citep{SemEval2018Task1} & 55.5   & 58.2                                      &  \bfseries  58.6                           \\
&   & \multicolumn{1}{c}{(\SI{+4.9}{\percent})} & \multicolumn{1}{c}{(\SI{+5.6}{\percent})}  \\
\bottomrule
\end{tabular}
}
\caption{The numerical results show that transfer learning can yield additional performance improvements based on an inductive knowledge transfer across \emph{tasks} (as opposed to the conventional strategy across \emph{datasets}). In our sent2affect method, the neural networks are first trained on a sentiment analysis dataset in order to learn an abstract representation of emotion-laden text, while the final dense layer is subsequently replaced and fine-tuned using the task-specific dataset. Performance is measured in terms of F1-score; \ie, the higher the better. The best-performing model for each dataset is highlighted in bold. The percentage changes refer to the relative improvement over the best baseline without transfer learning.}
	\label{tab:results_transfer_learning}
\end{table}

\section{Discussion} 
\label{sec:discussion}

\subsection{Comparison}

Our series of experiments reveals considerable and consistent performance improvements over default implementations of deep learning through the use of our customized networks. This points towards the need to customize deep neural networks according to the unique characteristics of the underlying task. 

In this paper, we refrained from evaluating performance on the basis of a single dataset and, instead, perform a holistic analysis, demonstrating that our customized networks outperformed the baselines in all experiments by up to \SI{23.2}{\percent}. Interestingly, our proposed modifications, such as with regard to regularization, were even able to learn the underlying relationships from the rather small datasets of merely 1,000 observations. However, we observe an overall pattern whereby the performance improvements tend to be higher when there is less language noise. In addition, we observe further improvement through the use of word embeddings, as these reduce the high-dimensional vectors with terms as one-hot encoding to lower-dimensional spaces. 

In the majority of experiments, the superior results stem from using a bidirectional LSTM as compared to a simple LSTM. We note that not only traditional machine learning but all network architectures required extensive training in order to ensure that embeddings and dropout layer functioned well together. Finally, the task of emotion recognition in affective computing is related to sentiment analysis, which infers a positive/negative polarity from linguistic materials. Hence, it is interesting to study whether one can further improve performance through an inductive transfer of knowledge from a different task (rather than a different dataset), despite the distinct objective, linguistic style, and annotation scheme. As a result, our sent2affect implementation of transfer learning establishes additional improvements of up to \SI{6.6}{\percent}.

\subsection{Deep-learning-based affect computing for decision support in social media}

As a proof of concept, we utilize our bidirectional LSTM to support the notoriously difficult task of classifying news into factual and non-factual. This demonstrates how affective computing can eventually facilitate decision support for social media platforms seeking to recognize and prevent the spread of \textquote{false news}. We utilize the dataset of \cite{Shu.2017} and predict whether a news item is factual. The prediction model is given by a logistic regression that is fed with the output of our affect prediction layer. Our approach achieves an accuracy of \SI{53.2}{\percent} when using the affective dimensions of the headlines and \SI{58.1}{\percent} when using separate affective dimensions of both headlines and content. This almost matches the reported baseline performance from prior research \citep{Rubin.2015}, where a content-based classifier was used to detect fabricated news items. However, we refrain from learning towards certain linguistic devices or individual stories. Instead, our approach ensures generalizability by identifying highly polarizing language as part of its decision support.

\subsection{Further use cases of deep-learning-based affective computing for better decision support}
\label{sec:applications}

Text-based affective computing drives decision support in a variety of application areas in which understanding the emotional state of individuals is crucial. \Cref{tab:applications} provides an overview of interesting examples from research, as well as actual use cases from businesses. This table is intended to give an overview of areas where decision support could potentially be improved through the use of our deep-learning-based models for affective computing. It is evident that affective computing facilitates decision-making in all operational areas of businesses, such as management, marketing, and finance. For instance, firms can infer the perceived emotion of customers from online product reviews and base managerial decisions on this data in order to support product development \citep{Ullah2016} and advertising \citep{Ang2000}. In a financial context, emotional media content has been identified as a driver in the decision-making of investors \citep{Prollochs2016}, which can thus serve as a decision rule for stock investments \citep{Gilbert2010}.

Beyond that, deep learning for emotion recognition could also facilitate public decision support with respect to politics and even education, as well as healthcare for individuals. For instance, affective computing can infer emotion concerning personal health conditions \citep{Anderson2011,Desmet2013,Greaves2013,VanDerZanden2014} and during learning processes \citep{Rodriguez2012}. Notably, all of the prior references engage in affect-aware decision-making, but have not yet evaluated the use of deep learning.  

\begin{table}[h]
	\centering
	\scriptsize
	\singlespacing
	\makebox[\textwidth]{%
		\begin{tabular}{p{2cm}p{3.8cm}p{7cm} c}
			\toprule
			\textbf{Domain} & \textbf{Application} & \textbf{Details} & \textbf{Reference} \\ 
			\midrule
			\emph{Management \&~marketing} &  Strategy development & Identification of perceived emotion towards products as a lever for product development & \citep{Ullah2016} \\[1.8em]
			& Brand management & Emotion analysis of firm-related tweets for reputation management & \citep{Al-Hajjar2015} \\[1.8em]
			& Churn prediction & Emotions within customer responses to marketing content serve as a predictor of purchase intention & \citep{Ang2000} \\[1.8em]
			& Preference learning & Examination of consumer behavior and emotional attitudes related to product preferences & \citep{Chitturi2007} \\ \midrule
			\emph{User \mbox{interaction}} 
			& Chatbots & Regulation of emotion of stranded passengers through chatbots & \citep{Medeiros.2017} \\[1.8em]
			& Social networks & Measurement of relationship strength in social networks with affective language as an indicator of emotional closeness & \citep{Marsden2012} \\ \midrule
			\emph{Finance}  & Investment decision & Prediction of stock market movements based on emotionally-charged content & \citep{Gilbert2010} \\[1.8em]
			& Economic growth indicator & Excitement and anxiety in media articles as indicators of financial stability and economic shifts & \citep{Nyman2015} \\ \midrule
			\emph{Politics} & Political participation & Emotion recognition for political participation and mobilization & \citep{Valentino2011} \\[1.8em]
			& Public monitoring & Hate speech detection on Twitter & \citep{Burnap2015} \\ \midrule
			\emph{Health} & Depression treatment & Analysis of emotional content for recognizing depressive symptoms in chat transcripts & \citep{VanDerZanden2014} \\[1.8em]
			& Suicide prevention & Early warning of suicide-related emotions in written notes & \citep{Desmet2013} \\[1.8em]
			& Public health forecast & Prediction of mortality from heart disease based on emotions expressed on Twitter & \citep{Eichstaedt2015} \\[1.8em]
			& Diagnosis & Emotional states as predictors of the willingness to disclose personal health information & \citep{Anderson2011} \\[1.8em]
			& Diagnosis & Social media emotion analysis for detecting poor healthcare conditions & \citep{Greaves2013} \\ \midrule
			\emph{Education} & E-learning & Improvement of learning experience through classifying and regulating e-learners' emotions & \citep{Rodriguez2012} \\[1.8em]
			\bottomrule
		\end{tabular}
	}
	\caption{Selected use cases in research and industry where deep-learning-based affective computing could help in improving decision support. Importantly, these works still rely upon traditional machine learning for emotion recognition and thus present viable opportunities for the use of our proposed deep learning framework.}
	\label{tab:applications}
\end{table}

\subsection{Implications for management and practice}

Even though deep learning has gained considerable traction lately, its use cases outside of academia remain scarce. A possible reason is located in the complexity of operationalizing deep neural networks. While recurrent architectures have previously been applied to sentiment analysis, the task of emotion recognition requires several modifications in order to obtain a better-than-random performance. This specifically applies to the proposed bidirectional processing of texts, regularization, and loss functions that can handle highly imbalanced datasets. As a direct recommendation for use cases of affective computing, we propose a shift towards customized network architectures, even for fairly small datasets of around 1,000 training samples, as in our case. Altogether, this highlights the need for a thorough understanding by practitioners of the available tools in order to benefit from deep learning. 

Affective computing for linguistic materials yields new opportunities for business models and consumer-centered services \citep{Li.2011, Doucet.2012,Dai.2015,Yin.2014}. Detecting and subsequently responding to the emotional states of users, customers, patients, and employees has the potential to significantly accelerate and improve management processes and optimize human-computer interactions. Here text remains a critical form of communication, while attempts have also been made to apply affective computing to speech or other multimodal input \citep{Calvo2010}, including visual data~\citep{Chen2017, ElAyadi2011, Shan2009}. Management should assess potential use cases in critical areas of operations from their own organizations. Our overview in \Cref{sec:background} provides illustrative examples, while further applications are likely to arise with recent methodological innovations. 

\subsection{Implications for research}

The process of improving the performance of affective computing would benefit considerably from a rigorous suite of baseline datasets. In the status quo, a variety of datasets with distinct goals and purposes is commonly used for benchmarking methodological innovations for affective computing. For instance, our literature survey identified four different strategies for annotating, including simple labels, multi-class labels, and numerical scores. Moreover, the set of affective dimensions varied between two (\ie, valence, arousal without explicitly naming emotions) and a set of 8 emotions (\eg, anger, disgust, surprise). However, this directly links to challenges concerning comparability and generalizability. In this sense, a network architecture that has been found effective for one annotation scheme might not work out for other datasets. On top of that, different labels prohibit transfer learning and thus impede performance. We therefore suggest a standardized approach to annotations. 

According to our literature review, datasets for affective computing vary in size from 1,000 instances to 7,902, and yet all of them remain fairly small when compared to other applications of deep learning. As a result, this is known to limit the performance of bidirectional LSTMs and other deep neural network architectures, which generally require large-scale datasets. For instance, datasets for sentiment analysis, such as the one used for our transfer learning approach, consist of up to \num[group-separator={,},group-minimum-digits=3]{100000} labeled samples. Future research should thus aim at creating larger datasets in order to enable the effective exploitation of deep learning.

\section{Conclusion}
\label{sec:conclusion}

Affective computing allows one to infer individual and collective emotional states from textual data and thus offers an anthropomorphic path for the provision of decision support. Even though deep learning has yielded considerable performance improvements for a variety of tasks in natural language processing, na{\"i}ve network architectures struggle with the task of emotion recognition. As a remedy, several modifications are presented in this paper: namely, bidirectional processing, dropout regularization, and weighted loss functions in order to cope with imbalances in the datasets.

Our computational experiments span categorical and dimensional emotion models, which require tailored algorithmic implementations involving, \eg, multi-class classification, as well as regression tasks and transfer learning. Our results show that pre-trained bidrectional LSTMs consistently outperform the baseline models from traditional machine learning. The performance improvements can even range up to \SI{23.2}{\percent} in F1-score for classification and \SI{11.6}{\percent} in MSE for regression. We propose sent2affect, a customized strategy of transfer learning that draws upon the different task of sentiment analysis (as opposed to different datasets, as is usually the case), which is responsible for further performance improvements of between \SI{5.6}{\percent} and \SI{6.6}{\percent}.

\section*{Acknowledgements}
\footnotesize
The authors gratefully acknowledge the financial support for Suzana Ili\'{c} from Prof. Kotaro Nakayama and Prof. Yutaka Matsuo, Graduate School of Engineering, The University of Tokyo, Tokyo, Japan.

\footnotesize
{\sloppy
	\begin{spacing}{1.35}
		\setlength{\bibsep}{3pt}
		\bibliographystyle{model5-names}
		\bibliography{AffectiveComputing}
	\end{spacing}
}
\normalsize
\appendix
\section{Recurrent Neural Networks}
This section presents our methods for inferring emotional states from narrative contents.

This is specifically grouped into classification tasks (where a set of emotions needs to be determined) and regression tasks (where the intensity of each affective dimension is represented by a numerical score).

We utilize a specific variant of the recurrent neural network, the long short-term memory model, which is known for being especially able to encode long dependency structures~\cite{hochreiter1997long}. The overall architecture is arranged according to three layers: (a)~an embedding layer that maps words in one-hot encoding onto low-dimensional vectors, (b)~a recurrent layer to pass information on between words, and (c)~a final dense layer for making the actual prediction. The latter varies according to whether it is an affective category or emotional intensity that is to be predicted. In the end, the weights in all neurons are estimated simultaneously during the training phase. The architecture of each layer is specified as follows:

\begin{enumerate}[(a)]
	\item \emph{Embedding layer}: Our first layer replaces the one-hot encoding of each word in the vocabulary with a numerical representation according to which words in terms of semantic meaning are optimized to have short distances between their word embeddings. For instance, the embedding of \emph{\textquote{good}} will eventually be closer to the word embedding of \emph{\textquote{great}} than to the word embedding of \emph{\textquote{boring}}. This includes explicit semantics and, in addition, the dense (as opposed to sparse) representation facilitates the optimization routines for training the subsequent layers. 
	\item \emph{Recurrent layer}: The word embeddings are then passed on to a recurrent layer, \ie a unidirectional LSTM or a bidirectional LSTM. The architecture of a recurrent layer is illustrated in \Cref{fig:RNN}. Here recurrent layers draw upon a single feedforward neural network $f$, for which the connections between neurons form cycles. As a result, recurrent layers can iterate over textual data word-by-word, thereby accumulating and memorizing information about the meaning of text in a hidden state vector. 
	
	\begin{figure}[!ht]
		\small
		\centering
		\includegraphics[scale=0.5]{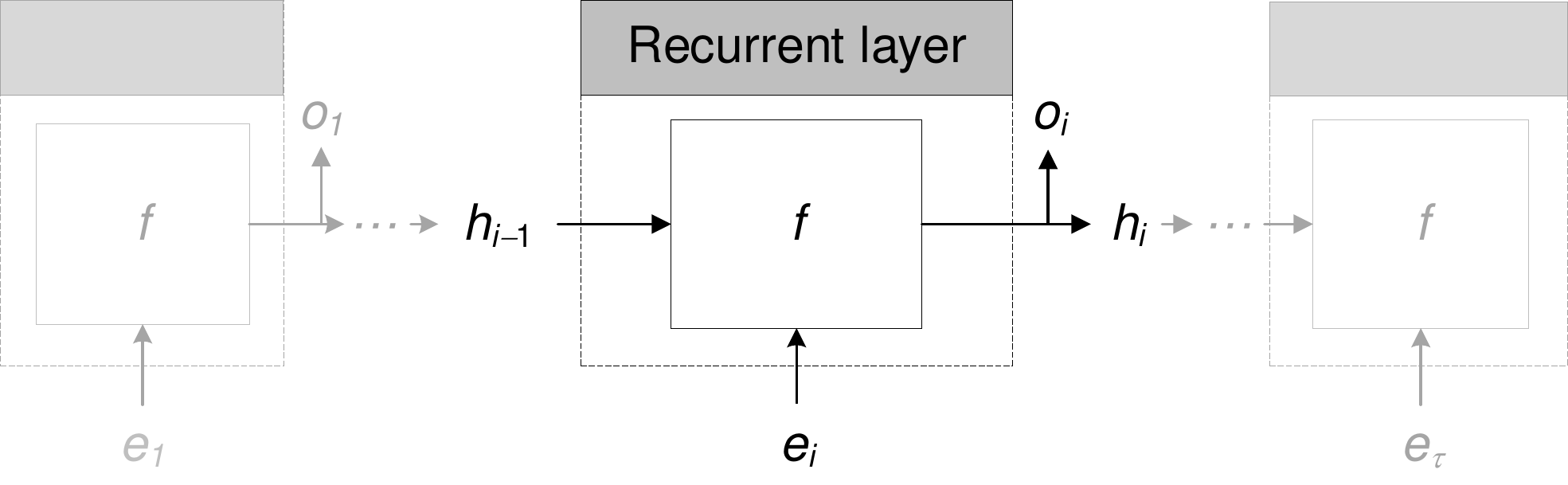}
		\caption{
			Schematic illustration of a recurrent layer that is unrolled over the input sequence. The $i$-th word is processed by feeding the embedding $e_i$ into the neural network $f$. This computes an output vector $o_i$ (that later links to the emotional state) and a hidden state $h_i$ that can pass information to the next, thereby encoding the sequence $e_1, \ldots, e_\tau$ in this hidden state vector.}
		\label{fig:RNN}
	\end{figure} 
	
	Formally, let $e_i$ be the word embedding of the $i$-th word. Furthermore, $f$ denotes a simple feedforward network that serves as the recurrent layer, while $h_i$ is a hidden state vector and $o_i$ when processing the $i$-th element in the sequence. When moving from term $i$ to $i+1$, the recurrent layer calculates the output $o_{i+1}$ through the neural network $f$ according to
	\begin{equation}
	o_{i+1} = f(h_i, e_{i+1}) .
	\end{equation}
	The recurrent layer is theoretically capable of accumulating text of arbitrary length, yet it requires a suitable design to overcome potential instabilities during optimization~\cite{bengio1994learning}. Therefore, this work follows common choices that advocate the use of long short-term memory networks. This architecture overcomes numerical instabilities by introducing an additional cell that stores the accumulated information with explicit update rules (see \Cref{fig:LSTM}). As an extension, we also experiment with a bidirectional variant (named BiLSTM) that duplicates the process in order to iterate over the word sequence in both directions. 
	
	\begin{figure}[H]
		\small
		\centering
		\includegraphics[width=1.0\textwidth]{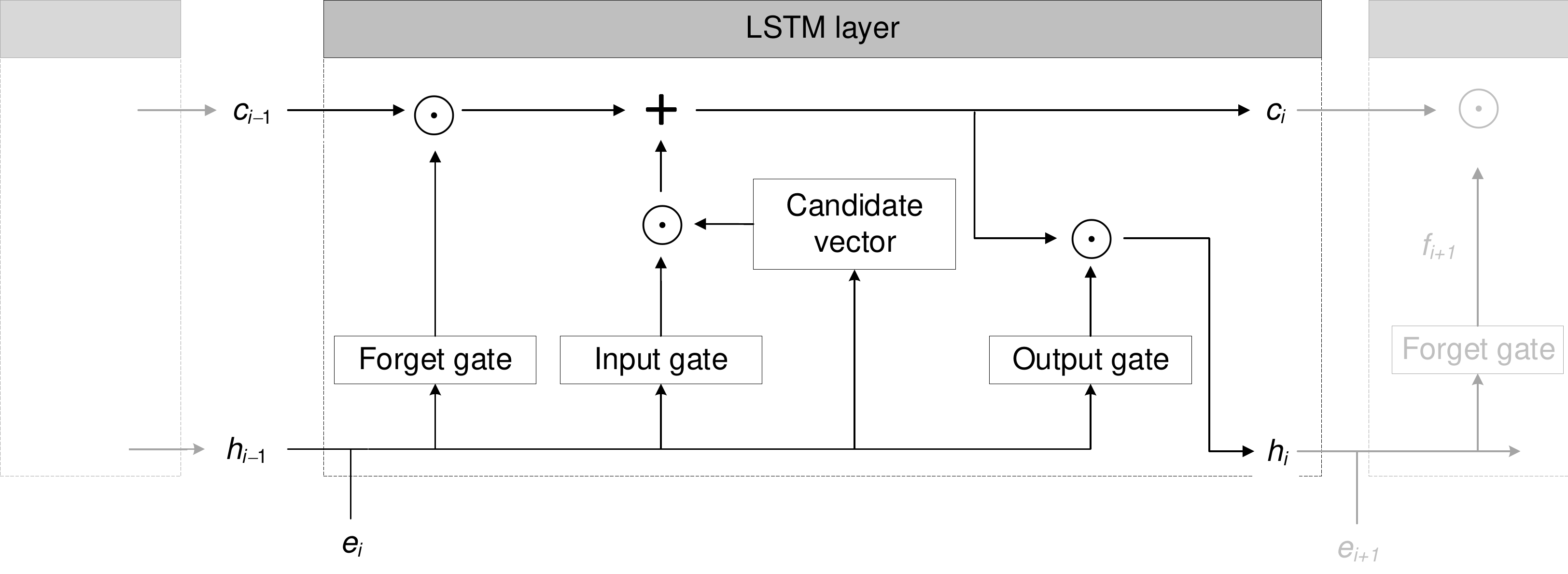}
		\caption{
			Schematic illustration of a long short-term memory that is again unrolled over the input sequence. The forget gate and the input gate are neural networks that update the cell based on the previous hidden state $h_{i-1}$, as well as the current input $e_i$. Furthermore, the output gate provides another neural network that computes the hidden state $h_i$. The hidden state $h_n$, belonging to the final word, then accumulates the complete document.}
		\label{fig:LSTM}
	\end{figure} 
	
	\item \emph{Dense layer}: The final dense layer $\psi$ draws upon the output of the LSTM layer with the aim of obtaining the final prediction output, \ie a label in a classification or a continuous score in a regression. 
\end{enumerate}

\subsection{Dense layer for affect prediction}

The choice of the dense layer for making the final prediction depends on the desired type, \ie whether we need to classify the document according to an emotional category or regress it against an intensity rating. Hence, the dense layer follows a linear operation in which every input neuron is connected to every output neuron through a coefficient that is optimized during training of the model. In general, dense layers are followed by activation functions, which are non-linear functions that increase the flexibility of the model or, in the case of a classification task, map the vector output from the LSTM layer onto a categorical representation. The choice of the activation function is governed by the underlying task and we discuss both in the following.

In the case of a classification, one commonly utilizes a softmax activation function $\sigma$, \ie a generalization of the logistic function that squashes its input values $x_1,\dots,x_k$ to values in the range $[0, 1]$. Mathematically, it computes
\begin{equation}
\sigma(x)_j = \frac{\exp{x_j}}{\sum_{k=1}^{n} \exp{x_k}},
\end{equation} 
for output $j$ with the additional property that $\sigma(x)_1,\dots,\sigma(x)_k$ sums to one. This allows us predict the membership with regard to $k$ different classes or categorical emotions by interpreting the estimate $\sigma(x)_j$ as a probability of $x$ belonging to a specific class. When only one class is desired, we compute $\argmax_{\kappa \in \{1,\dots,k\}} \sigma(x)_\kappa$ in order to identify the emotion with the highest probability.

In the case of the regression task, we implement an affine transformation $\alpha x^T + \beta$. Thereby, the underlying representation in the form of numerical values is aggregated onto a single numerical score that represents the intensity according to the desired affective dimension.

\end{document}